\documentclass[10pt,conference]{IEEEtran} 
\IEEEoverridecommandlockouts
\usepackage{cite}
\usepackage{amsmath,amssymb,amsfonts}
\usepackage{algorithmic}
\usepackage{graphicx}
\usepackage{textcomp}
\usepackage{xcolor}
\def\BibTeX{{\rm B\kern-.05em{\sc i\kern-.025em b}\kern-.08em
    T\kern-.1667em\lower.7ex\hbox{E}\kern-.125emX}}
\usepackage[ruled,linesnumbered]{algorithm2e}
\usepackage{makecell}
\usepackage{booktabs}

\usepackage{subcaption}
\usepackage{algorithmic}
\usepackage{graphicx}
\usepackage{textcomp}
\usepackage{xspace}
\usepackage{comment}
\usepackage{multirow}
\usepackage{tabu}
\usepackage{booktabs}
\usepackage{balance}
\usepackage{mathrsfs}
\usepackage{url}
\usepackage{hyperref}
\usepackage{cleveref}

\hypersetup{hidelinks,
	colorlinks=true,
	allcolors=black,
	pdfstartview=Fit,
	breaklinks=true,
 urlcolor=blue}

\newcommand{\ourname}{{InfeRE}\xspace} %NeRGic
\newcommand{\dataseta}{KB13\xspace}
\newcommand{\datasetb}{NL-RX-Turk\xspace}

\begin{document}

\title{\ourname: Step-by-Step Regex Generation via Chain of Inference
}

\author{\IEEEauthorblockN{Shuai Zhang, Xiaodong Gu, Yuting Chen, Beijun~Shen*}
\IEEEauthorblockA{\textit{School of Electronic Information and Electrical Engineering, Shanghai Jiao Tong University, Shanghai, China}\\
\{zhangshuai2000, xiaodong.gu, chenyt, bjshen\}@sjtu.edu.cn}}

\maketitle
\pagestyle{plain}

\begin{abstract}
 Automatically generating regular expressions (abbrev. regexes) from natural language description (NL2RE) has been an emerging research area. Prior studies treat regex as a linear sequence of tokens and generate the final expressions autoregressively in a single pass. They did not take into account the step-by-step internal text-matching processes behind the final results. This significantly hinders the efficacy and interpretability of regex generation by neural language models. In this paper, we propose a new paradigm called \ourname, which decomposes the generation of regexes into  chains of step-by-step inference. To enhance the robustness, we introduce a self-consistency decoding mechanism that ensembles multiple outputs sampled from different models. 
 % Experimental studies on two public benchmarks demonstrate that \ourname remarkably outperforms previous methods and achieves state-of-the-art performance. 
 We evaluate \ourname on two publicly available datasets, NL-RX-Turk and KB13, and compare the results with state-of-the-art approaches and the popular tree-based generation approach TRANX. Experimental results show that \ourname substantially outperforms previous baselines, yielding 16.3\% and 14.7\% improvement in DFA@5 accuracy on two datasets, respectively. %Particularly, \ourname outperforms the popular tree-based generation approach by 18.1\% and 11.3\% on both datasets, respectively, in terms of DFA@5 accuracy.

 %Our code and data are publicly available at {https://anonymous.4open.science/r/InfeRE-7DE0}.
\end{abstract}


\section{Introduction}
$\let\thefootnote\relax\footnotetext{* Beijun Shen is the corresponding author}$

 Regular expressions (abbrev. regexes) have been highly effective tools for text matching and manipulation, with extensive use in a broad range of fields. They enjoy robust support across nearly all programming languages~\cite{davis2019aren}, and are crucial to many practical applications such as performance testing tool \textit{JMeter}, command lines in Linux, and database operations that require powerful and flexible text processing. 
 Despite the wide utility of regexes, however, regexes can be complex, opaque, and challenging to compose, even for experienced programmers~\cite{luo2018marrying}. %\add{Therefore, automatic generation of regular expressions has attracted many attentions from academia and industry.}

Automatically generating regexes from natural language descriptions (NL2RE) has been an emerging research area~\cite{Semantic-Unify,Deep-Regex,SemRegex,SoftRegex,ye2020benchmarking,chen2020multi,ye2020optimal,DeepSketch,TransRegex}. Aside from earlier work on semantic parsing ~\cite{Semantic-Unify}, existing studies mainly treat NL2RE as a sequence-to-sequence learning problem and utilize an autoregressive language model. For example, Deep-Regex~\cite{Deep-Regex} employs an encoder-decoder model based on LSTM with attention. SemRegex extends the LSTM encoder-decoder model with reinforcement learning~\cite{SemRegex}, which rewards the model for generating diverse regexes. S$_2$RE~\cite{TransRegex} augments SemRegex by a new rewarding strategy to encourage the model to generate syntactically correct regexes. 

One common downside of existing studies is that they generate the entire regexes autoregressively. On the one side, autoregressive models generate regex one token at a time. Hence, they may not capture the real order of text processing behind the regexes. 
%resulting in unimodal distributions that cannot capture the full range of possible outputs. 
The generated tokens can violate the structure of regexes. 
Moreover, it is difficult to interpret the underlying processes and reasoning behind the generation. 

\begin{figure}[t]

\centerline{\includegraphics[width=\columnwidth, trim=90 120 110 110, clip]{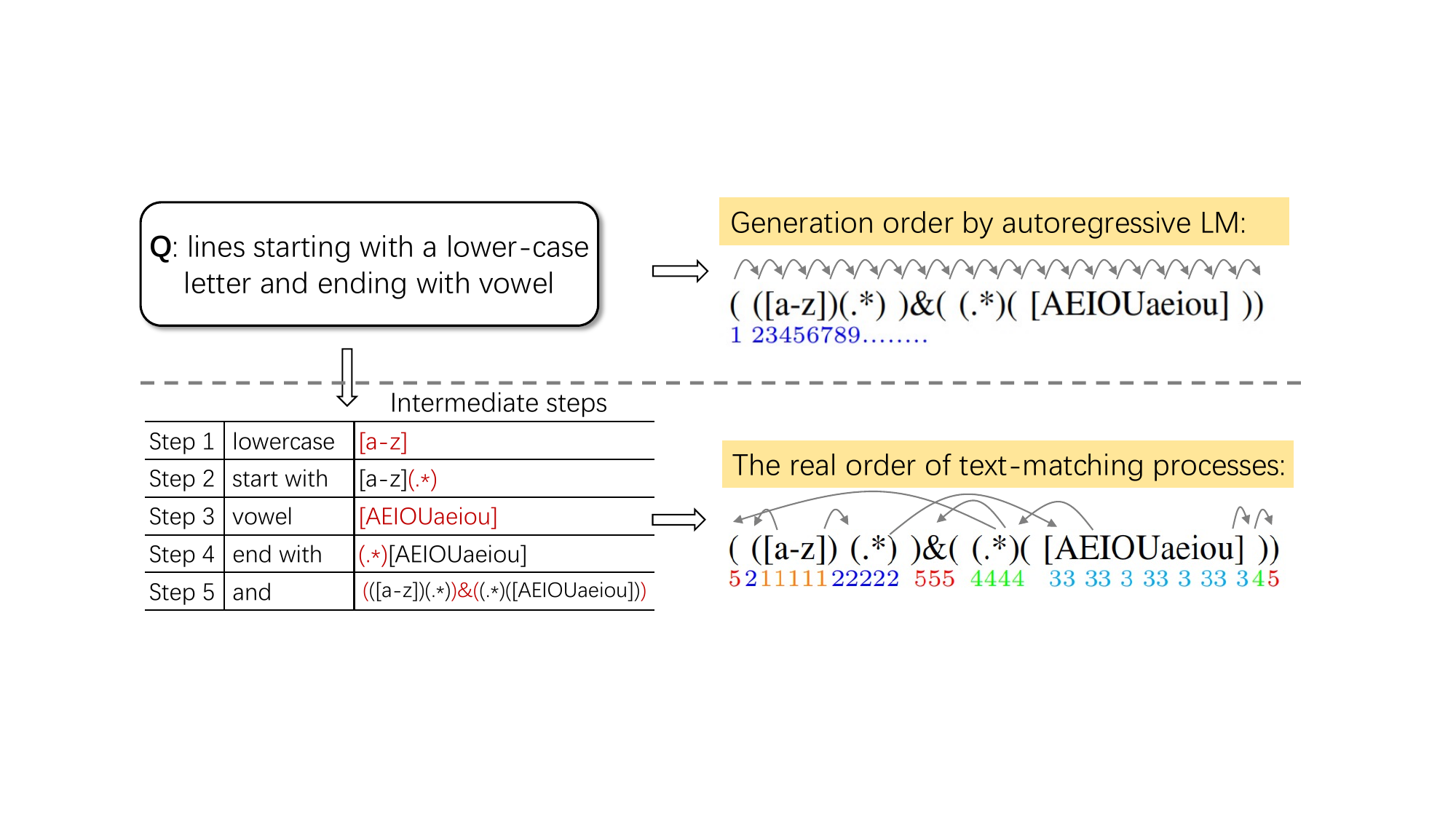}}
%\vspace{-2mm} 
    \caption{A motivation example of regex generation.}
\label{fig:motiveexample2}
%\vspace{-2mm} 
\end{figure}

%Neural inference chain represents the structural information of regexes more obviously since it mimics the process of building regexes step by step. 
 %For the query "\texttt{lines starting with a lower-case letter and ending with vowel}",  sequence-to-sequence language model recognizes the first two steps are ``\texttt{lines starting with a lower-case letter}'' and ``\texttt{ending with vowel}''. Then, language model generates corresponding internal steps \textbf{\texttt{step1=startwith(\textless low\textgreater)}} and \textbf{\texttt{step2=endwith(\textless vow\textgreater)}}. \textbf{\texttt{step3=and(step1,step2)}} means the combination of step1 and step2. Once all the steps are completed, automation script converts the chain of inference to plain regex \textbf{(([a-z])(.*))\&((.*)([AEIOUaeiou]))}.
Figure~\ref{fig:motiveexample2} shows an example of regex generation by conventional autoregressive models and the real text matching order. The integer numbers under the regex indicate the order for emitting each token. 
In conventional autoregressive models, the decoder generates tokens one by one based on previously decoded tokens. 
Right from the beginning, the decode must possess a comprehensive understanding of the entire output in order to give the correct quantity of left brackets before generating the genuinely semantically meaningful tokens.
% The decoder has to know about the whole output before emitting the first few tokens in order to give the correct left brackets\gu{not clear}.
In reality, the text-matching process does not follow a left-to-right order as the conventional sequence-to-sequence model assumes. For example, the first left bracket is emitted lastly when we consider the \texttt{and} operator. This example shows that conventional autoregressive models often fail to generate regex in the order of real text processing.

%\medskip\noindent\textbf{Our solution.}
Recent research~\cite{kojima2022large,wei2022chain} for large language models (LLMs) has shown that \emph{chain-of-thought prompting}, which decomposes a complex problem into a series of reasoning steps, can engage the reasoning capability of LLMs, resulting in substantial improvement in model performance. Additionally, multi-step generation allows for better interpretability of LLMs, since it provides understandable and transparent insights into how language model arrives at its predictions.
However, directly applying chain-of-thought prompting to the regex generation task is challenging, primarily due to the wide range of domains encompassed by existing LLMs, as well as the considerable expense associated with the manual formulation of prompts~\cite{DIN-SQL}. 

Inspired by the strengths of {chain-of-though} prompting, we propose \ourname, a novel formulation of regular expressions called \emph{chain-of-inference}, with each chain representing a sub-regex that indicates a specific text-matching procedure. Instead of generating plain regexes autoregressively, we formulate regex generation as step-by-step inference of internal processes for text matching. %\zs{Indeed, we generate steps all at ones by minimizing cross-entropy loss to reduce additional modifications to the language model.}%in analogy to the recently proposed \emph{chain-of-thought prompting}~\cite{kojima2022large,wei2022chain} for large language models. 

\ourname rationalizes the text-processing order through step-by-step generation. %For the example in Figure~\ref{fig:motiveexample2}, the first token generated by the conventional autoregressive model is the bracket corresponding to \texttt{and}, and the next token is the bracket corresponding to the \texttt{low-case letter}. There is a large information span between the first two tokens, which means that the model is required to have a clear overall understanding of the output at the early stage of decoding, undoubtedly increasing the difficulty of the problem. However, 
%\ourname follows the real order to gradually construct five sub-regexs, focusing on only specific information at each step, such as \texttt{lower-case letter} in step1 and  \texttt{vowel letter} in step3. Thus 
%\ourname decomposes 
Particularly, the original problem is decomposed into several simple subproblems by explicitly intermediate steps and the order in which these subproblems are addressed is consistent with the actual order of text processing. %First, we design a special representation for regexes in the form of chain of inference and develop an algorithm to automatically decompose plain regexes into chains of inference. 
As Figure~\ref{fig:motiveexample2}(b) shows, for the plain regex \texttt{(([a-z])(.*))\&((.*)([AEIOUaeiou]))}, the generation of a regex can be decomposed into 5 steps, with the step description (middle) and the internal results (right). Each step indicates a specific text-matching procedure, such as \texttt{step1=\textless low\textgreater} and \texttt{step2=startwith(\textless low\textgreater)}. Then, we train a language model to generate inference chains from natural language. 
Additionally, to enhance the robustness of model performance ~\cite{wang2022self}, we design a self-consistency decoding mechanism, aiming to ensemble multiple outputs and select the most consistent ones. 

We evaluate \ourname on two benchmark datasets, NL-RX-Turk and KB13, and compare the results with state-of-the-art approaches and the popular tree-based generation approach TRANX. Experimental results show that \ourname substantially outperforms previous baselines, yielding 16.3\% and 14.7\% improvement in DFA@5 accuracy on two datasets, respectively. Particularly, \ourname outperforms the popular tree-based generation approach by 18.1\% and 11.3\% on both datasets, respectively, in terms of DFA@5 accuracy. Ablation studies on components, backbone language models, and data sizes indicate that the proposed chain of inference and self-consistency decoding are effective and robust in various configurations. \ourname also demonstrates strong ability in low-resource regex generation. %Moreover, \ourname consistently outperforms SoftRegex and Deep-Regex baseline for different training set size.

In our paper, we mainly explore the application of \ourname to regex generation, but we believe that it is suggestive and may be broadly applied to other areas such as SQL generation, as they all follow a certain structural order.

The contributions of this work are as follows:
\begin{itemize}
\setlength\itemsep{0em}

\item We introduce \emph{chain-of-inference}, a novel representation format for regular expressions. Each chain indicates a specific text-matching procedure. %represents a sub-regex that 

\item We propose a novel paradigm of regex generation via chain of inference. 
%To our knowledge, we are the first to leverage the idea of \emph{chain of thought} for \zs{regex} %program 
%generation. 
In comparison with existing autoregressive approaches, \ourname builds sub-regexes in a manner that is similar to the human thought process. %\ourname complies with the real-matching processes of regexes and thus helps the language model better capture the structured information. More importantly, \ourname requires no architecture and loss function modifications.

\item We evaluate \ourname on two publicly available datasets and compare the results with state-of-the-art regex generation approaches as well as tree-based code generation methods. Experimental results show that \ourname substantially outperforms existing techniques on the datasets.
\end{itemize}

This paper is organized as follows: Section~\ref {sec:related} presents the related work. Section~\ref{sec:approach}  presents the algorithm and the implementation of two main mechanisms of \ourname, chain of inference and self-consistency decoding. Sections~\ref{sec:experiment} and V evaluate \ourname against state-of-the-art approaches on two benchmark datasets. %We further discuss the strengths and limitations in Section~\ref{sec:discuss}, and 
Section~\ref{sec:conclusion} concludes.

\begin{figure*}[t]
\centerline{\includegraphics[width=0.8\textwidth, trim=0 60 0 60, clip]{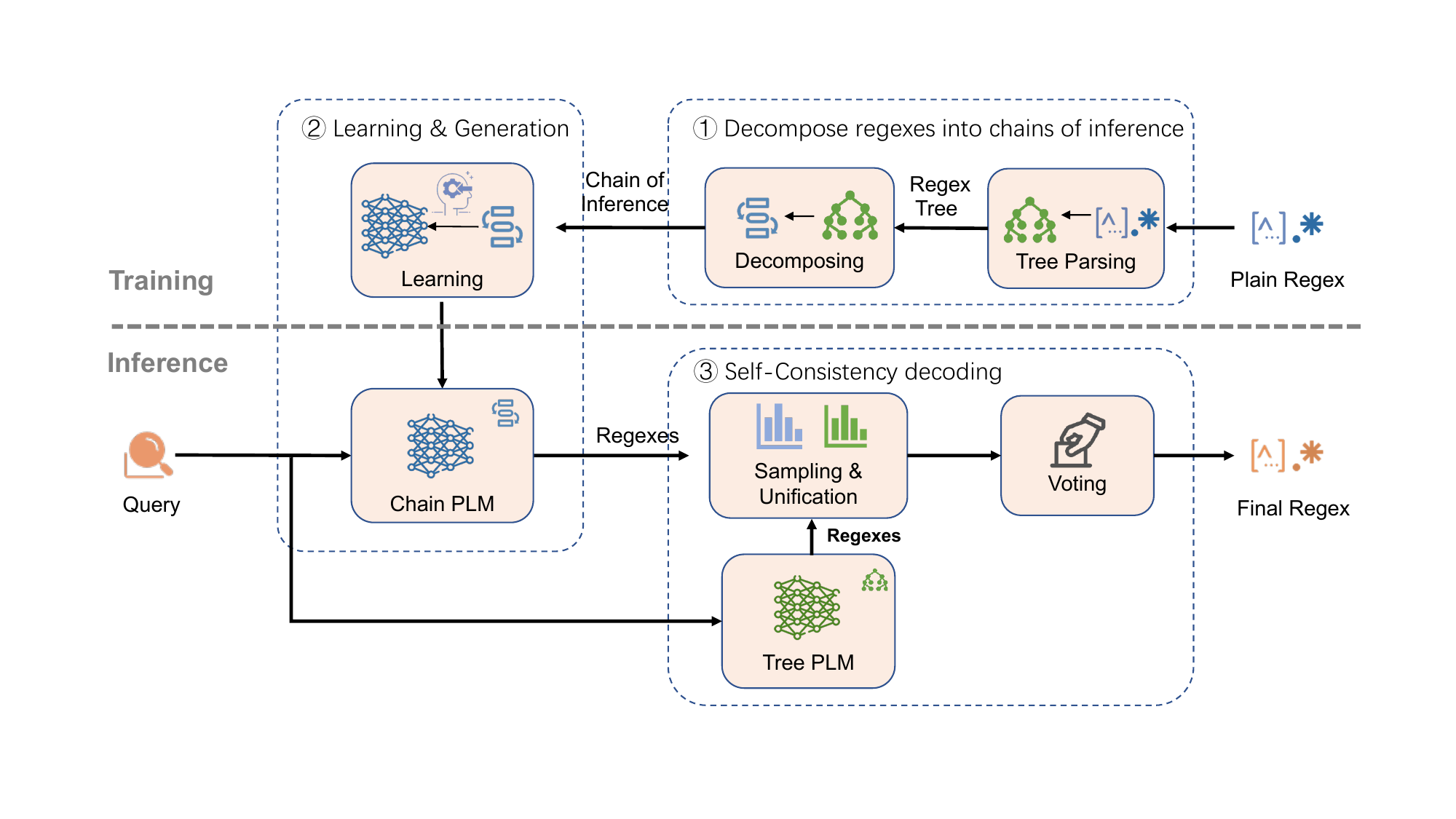}}
%\vspace{-2mm} 
    \caption{Overview of \ourname.}
    \label{fig:overview}
%\vspace{-2mm} 
\end{figure*}
\section{Related Work}
\label{sec:related}

\subsection{Automatic Regex Generation}
Automatically generating regexes has gained much attention due to the popularity and significance of regexes \cite{spishak2012type,liu2019lightweight,luo2018marrying}. %\add{Based on the forms of inputs, existing methods can be divided into three categories: %regex generation from examples, regex generation from natural language descriptions, regex generation from both natural language descriptions and examples.
%}
%\subsubsection{Regex Generation From Examples}
Early researches attempt to generate regexes from examples \cite{AlphaRegex,RegexGenerator,li2020flashregex,XML1,XML2,XML3}. They enumerate all possible regexes that are consistent with the provided examples and perform exhaustive searches with constraints. 
For example, AlphaRegex~\cite{AlphaRegex} synthesizes regexes by pruning out a large enumerative search space using over- and under-approximations of regexes. However, AlphaRegex can only generate simple regexes in specific forms. %from a set of positive and negative examples. 
RegexGenerator++~\cite{RegexGenerator} generates regexes based on genetic programming, but the correctness of generated regex is not guaranteed. FLASHREGEX~\cite{li2020flashregex} reduces the ambiguity of generated regexes by deterministic regex constraints. There is also a large number of studies that infer regexes from examples using XML schemas~\cite{XML1,XML2,XML3}.
A strong restriction of example-based approaches is that they rely heavily on high-quality examples. Hence, the systems tend to overfit when the examples provided by users are incorrect or limited.

%\textbf{Regex generation from natural language descriptions.} 
%\subsubsection{Regex Generation From NL descriptions}
Recently, there has been emerging research on NL2RE, i.e., regex generation from natural language descriptions. 
Ranta~\cite{Ranta1998} firstly developed a tool to convert NL queries to regexes based on abstract syntax rules. Kushman and Barzilay~\cite{Semantic-Unify} parsed natural language queries into regex by utilizing semantic unification. They also built the first benchmark and the DFA (deterministic finite automatons) metric for evaluating regex generation. 
Locascio et al.~\cite{Deep-Regex} proposed Deep-Regex, which is the first approach to employ deep learning for regex generation and provided another dataset of 10,000 natural language descriptions and regex pairs. 
Deep-Regex treats NL2RE as a machine translation task and employs an LSTM-based sequence-to-sequence model. SemRegex~\cite{SemRegex} alters the syntax-based training objective of Deep-Regex to a semantics-based objective by using reinforcement learning, which maximizes the DFA semantic correctness of generated regexes. 
SoftRegex introduces an additional reward model %EQ\_Reg 
to compute and soften the similarity of two regexes in SemRegex. 
%, which is a deep neural network for computing and softening the similarity of two regexes. 
Both SemRegex and Deep-Regex can generate correct regexes that may not resemble the ground truth benefiting from their reinforcement learning method. 
S$_2$RE~\cite{TransRegex} further improves the policy gradient to ensure the validity of generated regexes by rewarding the model if it generates valid regexes.

To the best of our knowledge, \ourname is the first attempt to employ pre-trained language models (PLM) in the field of automatic regex generation. Moreover, unlike current autoregressive models, \ourname first leverages the idea of \textit{chain of thought} for regex generation, which considers the real order of text processing and enhances the interpretability of large language models.

\subsection{Chain-of-Thought Prompting}

Recent research has shown that chain-of-thought prompting, i.e., prompting by step-by-step rationale, can significantly improve the performance of large language models on reasoning tasks, such as arithmetic, commonsense, and symbolic reasoning \cite{wei2022chain,narang2020wt5,wiegreffe2021reframing,wang2022rationale}. These approaches leverage a large language model to jointly generate the task output as well as the explanations to mimic the process of human reasoning. More importantly, chain-of-thought prompting requires no modifications to the architecture or training procedures of a large PLM. 

%Given the black-box nature of large language models, whether internal knowledge is being used properly is worth pondering. Previous studies have indicated that the spurious correlations learned from artifacts limits the performance of large language models. With this in mind, neural inference chain was first proposed to improve the interpretability of text-to-text models. Surprisingly, with the study deepening, researchers find that prompting models using explicit chain of thought can leverage models' internal knowledge and boost performance. Specifically, neural inference chain helps large language model decompose multi-step problems into intermediate steps, which means that models can allocate their computation resources more reasonably.

Related research reveals that the quality of chains of inference significantly affects the final model performance~\cite{ye2022unreliability}, and theirs effects vary in different downstream tasks~\cite{wei2022chain}. Wiegreffe et al. ~\cite{wiegreffe2021reframing} proposed an over-generation and filtration pipeline for producing model-generated rationales by utilizing GPT-3. They conducted a series of human studies and found that crowdworkers often prefer model-generated explanations to crowdsourced explanations. The follow-up study further indicates the sub-optimality of manually provided rationales increases sensitivity of large language models, and the ensemble of model-generated and human-written chains of inference can reduce this sensitivity \cite{wang2022rationale}. Therefore, appropriate chains of inference designed for specific downstream task are crucial.

Comparatively, \ourname adopts the idea of chain of thought and generates the step-by-step process of human reasoning instead of the final regex. However, \ourname is not a prompting method. That means, the generated sub-regexes are not taken as prompts to large language models. Instead, we leverage a large language model to generate chains of thought directly.

\section{Approach}
\label{sec:approach}
\subsection{Overview}
For a given natural language query $x$ = $x_1, ..., x_N$, the goal of NL2RE is to generate a regex $y$ = $y_1,...,y_T$ that aligns with the text transformation implied by the query. 
This problem can be realized using any sequence-to-sequence model, such as the Transformer~\cite{Transformer}. 
However, autoregressively generating the plain regex violates the real order of generation. 
Therefore, we design a novel \emph{chain-of-inference} paradigm for regex generation:
as opposed to directly generating $y$, we generate its chain of inference $s$ = $s_1$, ...,$s_K$, where each $s_t$ represents an internal text-matching process involved in the final regex. %\zs{ We will elaborate on the process of obtaining intermediate steps in Section ~\ref{learing and generation}}

Figure~\ref{fig:overview} shows the framework of our approach. The pipeline involves three main steps. First, we decompose plain regexes into chains of inference, namely, step-by-step operations in the regexes. Second, we train a sequence-to-sequence model to generate the chains of inference. These chains of inference are assembled into plain regexes. Third, we design a self-consistency decoding mechanism that ensembles multiple inference chains.

%\add{Compared with plain regexes, chain of inference describes the detailed construction process of regular expressions. Although structural information is also included in plain regexes, traditional autoregressive generation of the entire regexes does not follow the real order of text processing, and thus the structural information is not fully utilized in the generation process. Relatively, chain of inference makes the structural details of regexes more obvious by building sub-regex step by step and motivates language model to focus more on the specific process of constructing regular expressions}

%\add{Besides, the combination of chain of thought and self-consistency decoding helps \ourname maintain a robust performance. \ourname innovatively combine language models trained with data in different forms and consolidate the outputs of multiple PLMs to obtain the most consistent answer. On the one hand, the fine-tuning on chains of inference endows self-consistency decoding with a larger sampling space; on the other hand, self-consistency decoding reduces the randomness of inference results. The complementarity of the two mechanisms contributes to the performance of \ourname.}

\subsection{Decomposing Regexes into Chains of Inference}
\label{decomposing}

Our goal becomes how to convert a query $x$ into a chain of inference~$s$. A straightforward idea is to create a parallel corpus of ($x$, $s$) pairs and train a language model $p(s|x)$ on the corpus. However, our datasets are in the form of ($x$, $y$) pairs. It is impractical to label the chains of inference manually. Hence, we want to decompose the original regex $y$ into chains of processes $s$ automatically. 
The decomposition must satisfy two objectives: first, the decomposed chains must express the step-by-step text-matching process clearly; second, the chains cannot be redundant. Otherwise, the inferred chains may have extraordinary lengths, which degenerate the performance in turn. 

Our solution is simple: since regexes are tree-structured~\cite{DeepSketch} and the top-down hierarchy in the parsing trees explicitly delineates the step-by-step construction process of a regex~\cite{Deep-Regex}, we can decompose the original problem into small sub-problems by leveraging this inherent tree hierarchy. 

\begin{algorithm}[!t]
  \small
  \caption{Convert Plain Regexes to Chains of Inference} 
  \label{alg1} 
   \KwIn{$y$: original regex; $Op$: a dictionary of all operators and their corresponding number of operands}
            $NodeStack=(),C=(), B_{operand}=(), i=1$\;
            Parse $y$ into a tree $T(y)$\;
            Post-order traverse $T(y)$ and add nodes to $NodeStack$ in turn\;
            Reverse $NodeStack$\;

            \While{$NodeStack.size()\neq 1$}{
            $node=NodeStack.pop()$\;
	    \eIf{$node$ is operator}	{
     \tcc{Build one step in a chain of inference}
        $N_{node}=Op[node]$ //Lookup the operands 
        ~~~~~~~~~~~~~~~~~~~~~~~~~~~~~~~~~~~number for a node\\
    $s_{step}$="step" + $i$ + "=("\;
    \For {$i=0$ \textbf{to} $N_{node}$}{
        $o_i=B_{operand}.pop()$\;
        $s_{step}$ = $s_{step}$ + $o_i$ + ","\;
    }
    $s_{step}$ = $s_{step}$ + ")"\;
    $C.add(s_{step})$\;
    $B_{operand}.add($"step" + $i)$\;
    $i=i+1$\;
    }
    {
       $B_{operand}.add(node)$\;
    }
    }
    \KwOut{$C$: chain of inference}
\end{algorithm}

\begin{figure*}[t]
    \centerline{\includegraphics[width=0.92\textwidth, trim=20 0 30 0, clip]{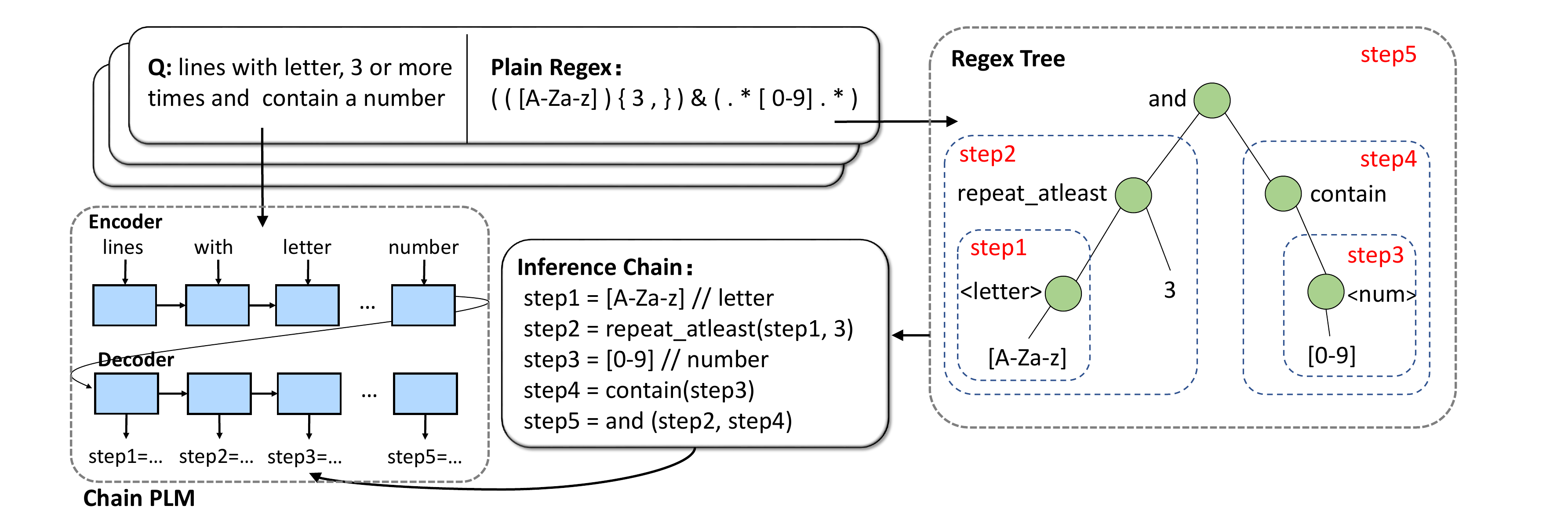}}
%\vspace{-2mm}     
    \caption{An illustration of creating chains of inference from plain regexes.}
    \label{fig:train}
%\vspace{-3mm} 
\end{figure*}

Algorithm \ref{alg1} describes the decomposition process. First, we parse the original plain regex $y$ into trees based on predefined rules defined in \cite{DeepSketch}. Nodes in the trees represent operators while edges represent the relationships between two sub-regexes. Table~\ref{table_dsl} lists all the rules for parsing plain regex into parenthetical notation for trees~\cite{ye2020benchmarking,DeepSketch}. The rules involve all operators and operands that appear in the benchmark datasets. Notably, the obtained parenthetical notation for trees can be readily mapped back to plain regex based on these established rules.

\begin{table}[!h]
\caption{Rules for parsing plain regex into parenthetical notation for trees~\cite{ye2020benchmarking,DeepSketch}.}
\label{table_dsl}
\small
\begin{center}
\setlength{\tabcolsep}{7mm}{
\begin{tabular}{l|c}
\toprule
\bf Operator r := &   \\

startwith ($r$) & $r.*$ \\
endwith($r$) & $.*r$ \\ 
contain($r$) & $.*r.*$ \\
not($r$)     & $\sim r$ \\
optional($r$) &  $r ?$ \\
star($r$)     &  $r*$ \\
concat($r_1$, $r_2$) & $r_1r_2$\\
and($r_1$, $r_2$) &  $r_1$ \& $r_2$ \\
or($r_1$, $r_2$) & $r_1$ | $r_2$ \\
rep($r$, $k$) & $r\{k\}$ \\
repeat\_least($r$, $k$) & $r\{k,\}$ \\
rep\_range($r$, $k_1$, $k_2$) & $r\{k_1,k_2\}$ \\
%\hline
\midrule
\bf Operand  := & \\
\textless let$\textgreater$ & [A-Za-z]\\
\textless cap$\textgreater$ & [A-Z]\\
\textless low$\textgreater$ & [a-z]\\
\textless num$\textgreater$ & [0-9]\\
\textless any$\textgreater$ & \textbf{.}\\
\textless spec$\textgreater$ & [-,;+:!@\#\_\$\%\&*=\^{}]\\
\bottomrule
\end{tabular}
}

\end{center}
\end{table}

Having obtained the parsing tree, we then traverse the tree in a post-order. Whenever we encounter an operator node, we regard the sub-regex corresponding to its sub-tree as a \textit{step} in the chain of inference. For the $i$-th node, we represent the sub-regex as step $i$ in the chain of inference. Then, we replace the sub-tree of the current operator with a \texttt{step-i} node. The size of the regex tree decreases continuously as the traversal proceeds. We repeat this process until there is only a single $step$ node left in the tree, which means the completion of the chain of inference.

Figure~\ref{fig:train} illustrates an example. For a pair of (query, regex) in the training set, training a sequence-to-sequence model to generate chain of inference is difficult. Hence, we parse the regex $y$=\texttt{(([A-Za-z])\{3,\})\&(.*[0-9].*)} into a tree and traverse the tree in post-order. When we encounter the first nonterminal node \texttt{\textless letter\textgreater} we create a sub-regex \texttt{step1=[A-Za-z]} in the chain of inference to indicate the first inferred matching process. The sub-tree rooted at \texttt{\textless letter\textgreater} is replaced with a special node \texttt{step1} to indicate the internal results after step 1. 
Next, we replace the sub-tree \texttt{repeat\_atleast(step1,3)} with a special node \texttt{step2} when we encounter the node \texttt{repeat\_atleast}.
Next, we replace the third occurred node \texttt{\textless num\textgreater} with a node \texttt{step 3} and append \texttt{step3=[0-9]} in the chain of inference.  %We pause the traversal at the operator node \texttt{repeat\_atleast}, which means the first step of inference,
%\textbf{\texttt{step1=repeat\_atleast(\textless letter\textgreater,3)}}, is finished. 
 %Similarly,  we obtain \textbf{\texttt{step2 = contain(\textless num\textgreater)} } next and the tree contains only three nodes currently: \texttt{step1}, \texttt{step2}, \texttt{and}. 
The traversal continues until meeting the root node \texttt{and}, where we generate \texttt{step5=and(step2,step4)} in the chain of inference and replace the sub-tree with \texttt{step5}.

%Different from previous research work on neural inference chain, we can obtain the neural inference chain of each regex through automatic script, while prior work relies on manual annotation entirely or partially, which is conducive to the application of our method on large datasets. In addition, prior work requires language model to generate the neural inference chain and the final answer together, while \ourname only needs to generate the neural inference chain alone and compute the final result utilizing script. Theoretically, the neural inference chain we propose can be extended to other tasks with tree structured outputs.

The decomposition step above is reminiscent of tree-based models in code generation such as TRANX~\cite{yin2018tranx}. Although both techniques generate code in a tree-based fashion, our architecture and training method are significantly different.
TRANX decomposes the tree into three types of actions and calculates their corresponding probabilities. This process deviates from the pre-training task of current PLMs, which makes it challenging to engage with large-scale language models.
% TRANX decomposes the tree into three types of actions and calculates their corresponding probabilities, a process that deviates from the pre-training task of current pre-training models, which makes it not easy to apply TRANX to large language models. 
In contrast, \ourname conforms to both the tree generation laws and the pre-training task of language models. Notably, \ourname does not require any modifications to the underlying model structure or training process, making it highly flexible and adaptable to various corpora and models.

\subsection{Learning and Generation}
\label{learing and generation}
Having parsed all regexes into chains of inference, we obtain a parallel corpus of (query, chain) pairs. 
We regard the generation of chains as a sequence-to-sequence learning task and leverage the Transformer model for the generation, where the encoder represents the input query as vectors while the decoder generates the target sequence (i.e., a chain of inferred regexes) based on the encoded hidden vectors. 

In order to leverage large domain knowledge on texts, we use T5 %(Text-to-Text Transfer Transformer)
~\cite{T5}, a popular PLM for sequence-to-sequence generation. T5 is first pre-trained on large corpora of texts using self-supervision. The pre-training enables T5 to gain rich generic knowledge, while the fine-tuning phase adapts the model to domain-specific tasks such as regex generation. 
Specifically, we take the open-source T5 model\footnote{https://github.com/google-research/text-to-text-transfer-transformer\#released-model-checkpoints} as the backbone of \ourname and further fine-tune it on the (query, chain) pairs as a text generation task by minimizing the cross-entropy loss. 

% We train the model by minimizing the cross-entropy loss: 
% \begin{equation}
% \theta^{*}=\arg \max _{\theta} \sum_{n=}^{N} \sum_{t=1}^{T_{n}} \log P\left(y_{t}^{n} \mid y_{\textless t}^{n}, x^{n}\right)
% \end{equation}
% where $N$ represents the batch size.

During the inference phase, we generate the chains of inference autoregressively by feeding the natural language query to the fine-tuned model. 
Then, we revert the generated chains to regex trees. We initialize the tree with the last step in the chain of inference, e.g., \texttt{step5=and(step2,step4)}. For each $step$-$i$ node, we replace the $step$-$i$ node in the sequence with its original sub-regex in the chain of inference. This process may introduce new $step$-$i$ node, so we repeat this process until there is no $step$ node left and get corresponding parenthetical notation for trees. Finally, we convert the parenthetical notation for trees into plain regex using the rules in Table~\ref{table_dsl}.
The whole procedures are summarized in Algorithm~\ref{alg2}.

\begin{algorithm}[t]
	\caption{Revert Chains of Inference to Plain Regexes} 
	\label{alg2}
		\KwIn{$C$: chain of inference; }
            $re=C.pop()$\;
            %\tcc{for example,we get $re=and(step1, step2)$}
            \While{`$step$-$i$' in $re$}{
                % get the value $i$ of $step$-$i$ node\;
                %\tcc{for example, i = 1}
                get the corresponding sub-regex $subre_i$ of $step$-$i$ from $C$[$i$]\;
                %\tcc{for example, step1=repeat\_atleast(\textless letter\textgreater,3)}
                $re$ = $re$.replace($step$-$i$, $subre_i$) \;
            }
            Convert $re$ to plain regex based on rules defined in Table~\ref{table_dsl}\;
    \KwOut{$re$: plain regex}
\end{algorithm}

\subsection{Self-Consistency Decoding}

\begin{figure*}[h]
\centerline{\includegraphics[width=1\textwidth, trim=0 160 0 160, clip]{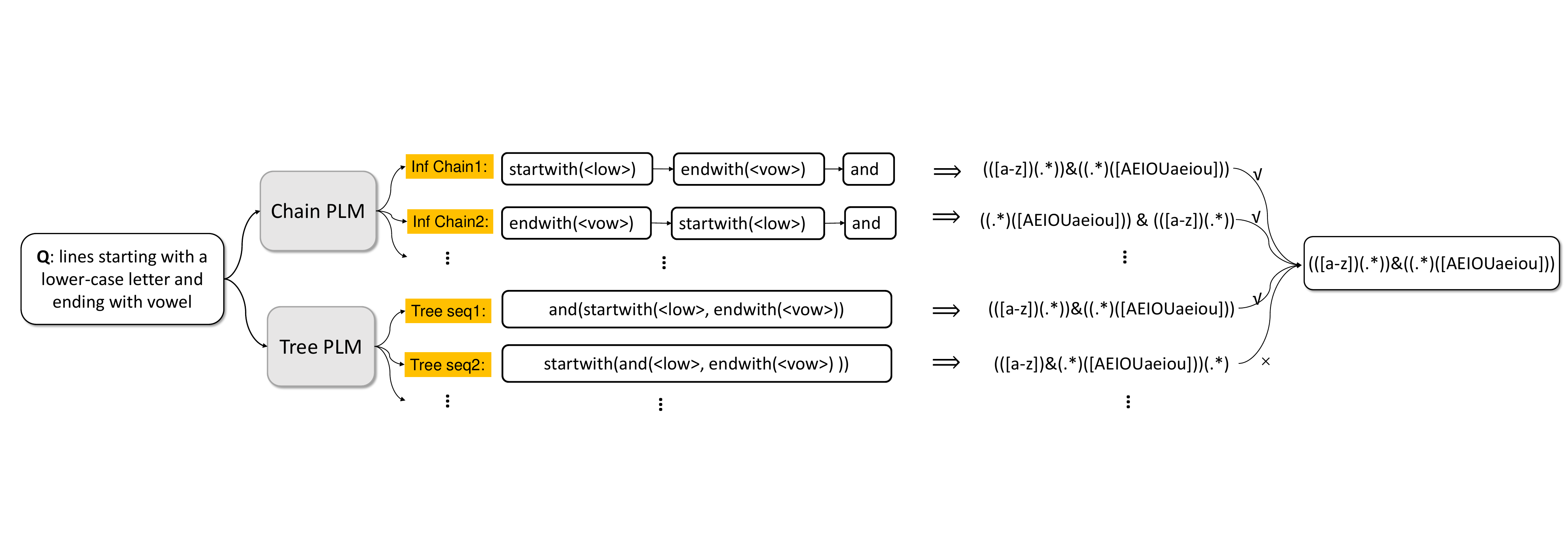}}
%\vspace{-2mm} 
    \caption{Illustration of self-consistency decoding.}
    \label{fig:self-consistency}
%\vspace{-2mm} 
\end{figure*}

Since we formulate regex generation as a chain of inference for text matching, it is natural to suppose that there are multiple chains that lead to the same text matching. %Different regexes may represent the same semantics, that is, multiple chains of thought may correspond to a single NL query. 
% We can simulate such process by sampling a group of neural inference chains from the language model's decoder.
Moreover, if multiple chains reach the same matching, that means the generated chains are more convincing~\cite{stanovich2000individual}. In other words, multiple wrong chains are unlikely to reach the same answer. Therefore, we can sample a group of inference chains from the language model, revert them to plain regexes, and select the most consistent ones (i.e., chains of inference that lead to the same regexes) as the final results.

Based on this idea, we propose a self-consistency decoding mechanism that selects the most consistent answer by sampling in the output space of language model~\cite{wang2022self}. 
Figure~\ref{fig:self-consistency} illustrates the process of self-consistency decoding. 

First, we employ two language models: a chain PLM which is trained on (x, s) pairs to generate chains of inference, and a tree PLM which is trained on (x, y) pairs to generate the parsing trees in parenthetical notation (e.g., \texttt{and(startwith(\textless low\textgreater, endwith(\textless vow\textgreater))}). 

At the decoding phase, each language model samples $k$ outputs. The total $2k$ outputs are reverted to plain regexes and are used to select the final regex using a plurality vote strategy. Since different regexes may represent the same semantic, we take a plurality vote based on semantic consistency.\footnote{We measure the semantic consistency using the DFA-EQ metric introduced in Section ~\ref{metric}.} One output represents one vote. Regexes with the same semantics will share the votes, and we randomly select one of them as the representative. Finally, \ourname chooses the regex that receives the most votes as the final output. If there is more than one answer with equal votes, we take the answer first encountered.

%~\gu{is it possible that none of the 2k outputs will be selected?}\zs{no. add an explanation}. 
Compared with standard decoding methods, self-consistency encourages correctness and diversity at the same time~\cite{wang2022self}.

% \ourname marginalizes over the chains and integrates the outputs from different decoders together\gu{how to marginalize? how to integrate?} before taking a plurality vote~\gu{what is plurality vote? how is it computed?}. Since different regexes may represent the same semantics, the voting is based on semantic consistency. That means, regexes with the same semantics\gu{how to measure the semantic consistency?} share the votes. Finally, \ourname chooses the most consistent regexes as the final answer. 

It is worth noting that our idea is different from previous studies \cite{wang2022self,wang2022rationale} on diversity driven decoding, i.e., introducing randomness in the input and output space of a single model.  %Unlike previous work which introduce randomness into a single model, 
We train multiple language models to generate both regexes and chains of inference. The different forms of output can greatly enrich the input and output space of language models and thus enhance the self-consistency decoding mechanism.

\section{Experimental Setup}
\label{sec:experiment}

\subsection{Research Questions}
We evaluate \ourname by addressing the following research questions:
\begin{itemize}
\item \textbf{RQ1: How effective is \ourname in generating regexes from natural language descriptions?} 
%To answer \textbf{RQ1}, 
We evaluate the performance of \ourname on two widely used datasets and compare it against state-of-the-art regex generation approaches and a tree-based code generation approach.

\item \textbf{RQ2: What is the effect of chain of inference?} 
%To answer \textbf{RQ2}, 
We conduct an ablation study on the efficacy of the chain of inference. We further apply chain of inference to different PLMs.

\item \textbf{RQ3: What is the effect of self-consistency decoding and how the number of output samples affects the performance of self-consistency decoding?}  We conduct an ablation study to analyze the effect of self-consistency decoding mechanism. As the self-consistency decoding relies heavily on the number $k$ of output samples, we further vary the number $k$ to explore the performance of self-consistency decoding.
% \item \textbf{RQ3+RQ4: What is the effect of self-consistency decoding?} 
% %To answer \textbf{RQ3}, 
% We conduct an ablation study to analyze the effect of the self-consistency decoding mechanism.

% \item \textbf{RQ4: How does the number of output samples affect the performance of self-consistency decoding?} 
% %To answer \textbf{RQ4}, 
% We further vary the number $k$ to explore the performance of self-consistency decoding, as self-consistency decoding relies heavily on the number $k$ of output samples.

\item \textbf{RQ4: How does different backbone PLMs impact the performance of \ourname?} 
%To answer \textbf{RQ4}, 
We study the effect of different backbone PLMs on performance. We replace the backbone with popular PLMs of different types and sizes and compare their performance.

\item \textbf{RQ5: What is the impact of data size on the performance of regex generation?} 
In practical applications, sufficient high-quality parallel corpora are difficult to obtain. To answer this question, we aim to study the effect of data size on performance. We vary the size of the training set and compare the performance of various approaches under different data sizes.

%\gu{RQ3-RQ4 need to be merged. New RQ2: What is the effect of chain of inference and self-consistency? New RQ3: How do different configurations affect the performance? }

% \item \textbf{RQ4: How do different hyperparameters affect the performance of \ourname?} .
\end{itemize}

\subsection{Comparison Methods}
We compare \ourname with five regex generation methods, including the state-of-the-art method S$_2$RE~\cite{TransRegex}. Since S$_2$RE is not pre-trained, we extend S$_2$RE to S$_2$RE-T5 by replacing the LSTM backbone with T5. \ourname is also related to tree-based generation models. Therefore, we compare it with TRANX~\cite{yin2018tranx}, a popular tree-based code generation method. Specifically, we compare \ourname with the following baselines.

\begin{itemize}
\setlength\itemsep{0em}
\item \textbf{Semantic-Unify}~\cite{Semantic-Unify}: a grammar-based approach that directly parses natural language description into regexes. Semantic-Unify learns a combinatory categorial grammar, which consists of words or phrases with their corresponding regular expression functions.

\item \textbf{Deep-Regex}~\cite{Deep-Regex}: an LSTM sequence-to-sequence model with attention. Deep-Regex regards regex generation as a 1-to-1 translation task without considering semantically equivalent regexes with different forms. 
The model can be trained by either maximizing likelihood (Deep-Regex$^{\mathrm{MLE}}$~\cite{Deep-Regex}) or maximizing marginal likelihood (Deep-Regex$^{\mathrm{MML}}$~\cite{SoftRegex}).

\item \textbf{SemRegex}~\cite{SemRegex}: the same model as Deep-Regex, but trained using policy gradient~\cite{williams1992simple} to reward the model for generating equivalent regexes in different forms.

\item \textbf{SoftRegex}~\cite{SoftRegex}: an extension of SemRegex by training an additional reward model (EQ\_Reg), which measures the equivalence probability of two regexes and then uses the probability to reward the sequence-to-sequence model.

\item \textbf{S$_2$RE}~\cite{TransRegex}: an LSTM sequence-to-sequence model trained via policy gradient. S$_2$RE introduces both syntactic and semantic validity to generate more valid regexes. To our knowledge, S$_2$RE is the state-of-the-art approach for regex generation. 

\item \textbf{S$_2$RE-T5}~\cite{TransRegex}: a variant of S$_2$RE by replacing LSTM with T5, which is a PLM with a larger number of parameters. For fair comparison, we implement S$_2$RE-T5 using the  configurations in ~\cite{TransRegex}. 
%S$_2$RE-T5 also introduces both syntactic and semantic validity to reward the sequence-to-sequence model.

\item \textbf{TRANX}~\cite{yin2018tranx}: a tree-based code generation model that is widely used in recent studies~\cite{jiang2021ast,dong2022antecedent,beau2022impact,shen2022incorporating}. Similar to \ourname, TRANX decouples code generation as a series of tree construction actions using user-defined, task-dependent grammar.
\end{itemize}

We reproduce the results of Deep-Regex, SoftRegex, and TRANX using their officially published code. 
For S$_2$RE-T5, we initialize the parameters by pre-training the model on the (query,regex) pairs training set and leverage policy gradient to train
the model with the semantics-based and syntactic-based objective.
For baselines that are not open-source, we directly excerpt the statistics from the original paper and leave blanks if the results are not reported.

\subsection{Evaluation Metrics}
\label{metric}

We measure the performance of regex generation using two commonly used metrics:

\smallskip \noindent \textbf{DFA-EQ}~\cite{Semantic-Unify}.  
Since regexes in different forms may represent the same semantic, DFA-EQ converts regexes to deterministic finite automatons (DFAs) and measures their semantic equivalence by comparing their DFAs. If their DFAs exactly match, the two regexes are semantically equivalent. 
DFA-EQ@m measures the proportion of cases for which the top-$m$ generated regexes contain at least one regex that is semantically equivalent to the ground truth. In our experiments, we compute the top-m DFA-EQ accuracy with m = 1 and 5.%according to their DFAs. 
%DFA-EQ measures how many regexes generated are semantically equivalent to the ground truth according to their DFAs. 

\smallskip \noindent \textbf{EM} (Exact Match).
EM measures the ratio of regexes that exactly match ground-truth regexes.  Different from the DFA-EQ, EM compares two regexes in plain texts.

\subsection{Datasets}

We evaluate \ourname on two widely used datasets, namely, \datasetb~\cite{Deep-Regex} and \dataseta~\cite{Semantic-Unify}. %Since our research task is NL2RE, we discard positive and negative examples in the datasets.
The statistics of the two datasets are presented in Table~\ref{table_data}. 

\begin{table}[t]
\caption{Statistics of the datasets.}
\label{table_data}
%\small
\begin{center}
\setlength{\tabcolsep}{4mm}{
\begin{tabular}{l|ccc}
\toprule
\bf Dataset & \bf \ Train &\bf Dev & \bf Test \\
\hline
\datasetb & 6,500 & 1,000 & 2,500 \\ 
\dataseta & 618 & 206 & 206 \\
\bottomrule
\end{tabular}
}
%\vspace{-1mm} 
\end{center}
%\vspace{-5mm} 
\end{table}
\begin{table*}[t]
\caption{Performance of various approaches on regex generation %on two benchmark datasets. 
(SC=Self-Consistency; CI=Chain of Inference).}
\label{table_rq1}
\begin{center}
%\small
\setlength{\tabcolsep}{2.1mm}{
\begin{tabular}{lcccccc}
 \toprule
        \multirow{2}{*}{\textbf{~Approach}} &  \multicolumn{3}{c}{\textbf{\datasetb}}
        & \multicolumn{3}{c}{\textbf{\dataseta}}  \\
        \cmidrule(r){2-4} \cmidrule(r){5-7}  
& \textbf{DFA-EQ@1(\%)} & \textbf{DFA-EQ@5(\%)} & \textbf{EM(\%)}  & \textbf{DFA-EQ@1(\%)} & \textbf{DFA-EQ@5(\%)} & \textbf{EM(\%)} \\
        \midrule
~Semantic-Unify & 38.6 & ---& --- & 65.5  & --- & ---  \\
~Deep-Regex$^{\mathrm{MLE}}$~~~ & 60.3 & 76.0 & 40.7 &  66.5 & 75.7  & 55.8   \\
~Deep-Regex$^{\mathrm{MML}}$ & 62.4 & 76.8  & 39.2 &  68.2 & 77.7& 56.8 
  \\ 
~SemRegex   & 62.3 & ---& --- & 78.2 & ---& ---   \\ 
~SoftRegex  & 62.8 & 72.1 & 41.5  & 78.2 & 79.6 & 62.1   \\ 
~S$_2$RE &  62.8 & --- & ---  & 78.2 & --- & ---  \\ 
%\hline
~{S$_2$RE-T5} &67.6 & 85.7  & 54.4 &  82.0 & 88.8 & 71.4  \\ 
%\hline
~TRANX &  58.8 & 75.6 & 44.0  & 73.8 & 82.0 & 61.2  \\ 
\hline
~\ourname (ours) & \bf 69.2 & \bf 89.3  & 53.4  & \bf 82.5 & \bf 91.3 & 69.4  \\ 
 ~~~~- $w/o$ SC  & 67.8 & 85.9 & \bf 55.5 & 81.6  & 87.9 & \bf 72.3   \\
 ~~~~- $w/o$ SC+CI & 67.2 & 85.5 & 54.2 & 82.0 & 88.8  & 70.4   \\

\bottomrule
%\multicolumn{7}{l}{$^{\mathrm{*}}$ SC=Self-consistency; CI=chain of inference.}
\end{tabular}
}
\end{center}
\end{table*}

\datasetb is the largest dataset which contains 10,000 regexes with natural language annotations. Due to the heavy workload of manual labeling, Locascio et al. employ a generate-and-paraphrase process for annotating data automatically~\cite{Deep-Regex}. They initially generated 10,000 regexes with natural language descriptions stochastically based on manually-crafted grammar. Then, human annotators were asked to paraphrase the synthetic natural language descriptions. 

\dataseta is a smaller and easier dataset comprising 824 (query, regex) pairs. The dataset was created by crowdsourcing: the authors first asked crowdsource workers to write natural language queries. Then, each query was presented to experienced programmers who wrote the corresponding regexes. On average, the length of natural language descriptions in \datasetb is 50\% longer than \dataseta.
Following previous studies~\cite{Deep-Regex,SemRegex,SoftRegex,TransRegex}, we split the original dataset into train, develop, and test sets. 

\subsection{Implementation Details}
We build our models on top of the T5 using the same configuration as T5-base (d\_ff $=3072$, d\_kv $=64$, d\_model $=768$). Both the encoder and decoder consist of 12 blocks with a dropout probability of 0.1. On both datasets, the batch size is set to 12 per GPU. We optimize the model using Adam optimizer with a learning rate of 3e$^{-5}$ and an epsilon of 1e$^{-8}$. For the self-consistency decoding, following previous studies~\cite{wang2022self}, we set the number $k$ of outputs sampled from each model to 40. 

We train the models for 30 and 60 epochs on \datasetb and \dataseta, respectively. The checkpoint that obtains the best performance on the development set is selected for testing. All experiments are carried out on a workstation with 2 RTX 2080Ti GPUs and an operating system of Ubuntu 18.04.4
LTS. %(CUDA Version 10.2). 

\section{Results}
\label{sec:results}
\subsection{Effectiveness of \ourname in Regex Generation (RQ1)}
% \begin{table*}[t]
% \begin{center}
% \small
% \setlength{\tabcolsep}{1.4mm}{
% \begin{tabular}{l|ccc|ccc}
%  \toprule
%         \multirow{2}{*}{\textbf{Approach}} &  \multicolumn{3}{c|}{\textbf{\dataseta}}
%         & \multicolumn{3}{c}{\textbf{\datasetb}}  \\
%         \cmidrule(r){2-4} \cmidrule(r){5-7}  
% & \textbf{DFA-EQ@1} & \textbf{DFA-EQ@5} & \textbf{EM}  & \textbf{DFA-EQ@1} & \textbf{DFA-EQ@5} & \textbf{EM} \\
%         \midrule
% Semantic-Unify & 65.5\%  & --- & ---  & 38.6\% & ---& ---\\
% Deep-Regex$^{\mathrm{MLE}}$ &  66.5\% & 75.7\%  & 55.8\%  & 60.3\% & 76.0\% & 40.7\% \\
% Deep-Regex$^{\mathrm{MML}}$ &  68.2\% & 77.7\%& 56.8\% 
%   & 62.4\% & 76.8\%  & 39.2\%\\ 
% SemRegex   & 78.2\% & ---& ---  & 62.3\% & ---& --- \\ 
% SoftRegex    & 78.2\% & 79.6\% & 62.1\%  & 62.8\% & 72.1\% & 41.5\% \\ 
% S$_2$RE   & 78.2\% & --- & ---  &  62.8\% & --- & ---\\ 
% \hline
% \ourname (ours)   & \bf 82.5\% & \bf 91.3\% & 69.4\%  & \bf 69.2\% & \bf 89.3\%  & 53.4\%\\ 
%  ~~- $w/o$ SC   & 81.6\%  & 87.9\% & \bf 72.3\%   & 67.8\% & 85.9\% & \bf 55.5\%\\
%  ~~- $w/o$ CI  & 82.0\% & 88.8\%  & 70.4\%   & 67.2\% & 85.5\% & 54.2\%\\

% \bottomrule
% %\multicolumn{7}{l}{$^{\mathrm{*}}$ SC=Self-consistency; CI=Chain of Inference.}
% \end{tabular}
% }
% \caption{Performance of various approaches on regex generation %on two benchmark datasets. 
% (SC=Self-Consistency; CI=chain of inference).}
% \label{table_rq1}
% \vspace{-1mm} 
% \end{center}
% \end{table*}

Table~\ref{table_rq1} compares the performance of various methods on the two datasets.
Overall, \ourname achieves the best performance among all the approaches. 

Compared with state-of-the-art NL2RE methods, \ourname demonstrates strong strength,
%  the top-5 DFA-EQ accuracy of InfeRE is about 13\% and 12\% greater than that of Deep-Regex$^{\mathrm{MML}}$ and SoftRegex on NL-RX-Turk and KB13, respectively
with the top-5 DFA-EQ accuracy increased by 16.3\% and 14.7\% against S$_2$RE on \datasetb and \dataseta, respectively. 
A more significant improvement can be observed in the EM metric, which is increased by 33.7\% and 16.4\% on two datasets compared to S$_2$RE.

We notice that \ourname also surpasses S$_2$RE-T5, a PLM with the same number of parameters as \ourname. The DFA@1 and DFA@5 are improved by 2.3\% and 4.2\% on \datasetb. The results indicate that the improvement does not solely come from pre-training, which affirms the effectiveness of the proposed chain of inference and self-consistency decoding mechanisms.

Compared with the tree-based code generation model TRANX, \ourname gains 13.2\%, 14.7\%, and 22.1\% improvement in the three metrics, respectively, demonstrating the superiority of \ourname over the traditional tree-based method.

One notable phenomenon is that the improvement is more significant on \datasetb than on \dataseta. One main reason is that \datasetb is larger and more complex than \dataseta,  allowing the model to capture richer knowledge from the chains of inference.

%\subsection{Qualitative Analysis}
%To demonstrate the effectiveness of our approach, 

\begin{table*}[!t]
\caption{Generated regex examples by various approaches. The middle column shows the final regexes. For ease of analysis, we also present the intermediate trees in a parenthetical notation in the right column. All results are generated via greedy decoding except \ourname with self-consistency decoding.}
% \vspace{-2mm}
\label{table_examples}
\centering
%\small
\setlength{\tabcolsep}{4mm}{
\begin{tabular}{lll}
\toprule
\textbf{Example 1.}& \multicolumn{2}{l}{Query: \textit{either start with string \textless M0$\textgreater$ or end with any number.}} \\
\hline
\textbf{Gold:} & \textbf{((.*)([0-9]))|((\textless m0\textgreater)(.*))} & \textbf{or(endwith(\textless num\textgreater),startwith(\textless m0\textgreater))}\\
\textbf{Deep-Regex$^{MML}$:} & ((\textless m0\textgreater)|((.*)([0-9])))(.*)& \textcolor{red}{startwith(or(\textless m0\textgreater},endwith(\textless num\textgreater)))\\
    \textbf{SoftRegex:} & ((\textless m0\textgreater)|(.*)([0-9]))(.*)&\textcolor{red}{startwith(or(\textless m0\textgreater},endwith(\textless num\textgreater)))\\
\textbf{\ourname -w/o SC+CI:} & ((\textless m0\textgreater)|((.*)([0-9])))(.*)&\textcolor{red}{startwith(or(\textless m0\textgreater},endwith(\textless num\textgreater))) \\

% \red{startwith(or(\textless m0\textgreater,endwith(\textless num\textgreater)))}
% \colorbox{gray}{startwith(or(\textless m0\textgreater,endwith(\textless num\textgreater)))}\\
\textbf{\ourname -w/o SC:}& ((\textless m0\textgreater)(.*))|((.*)([0-9])) &  \textcolor{green!60!black}{or(startwith(\textless m0\textgreater),endwith(\textless num\textgreater))}\\
% \colorbox[RGB]{225,225,225}{\makecell[l]{step1=\textless m0$\textgreater$ \\
% step2=\textless num$\textgreater$ \\
% step3=startwith(\textless m0\textgreater) \\
% step4=endwith(\textless num\textgreater) \\
% % \ \ \ \ \ \ \ \ \ \ \ \ \ \ \ \ \ \ \ \ \ \ \ \ \ \  
% step5=or(step3,step4) \\
% \textbf{result}=((\textless m0\textgreater)(.*))|((.*)([0-9]))\\}} \\
% \ \ \ \ \ \ \ \ \ \ \ \ \ \ \ \ \ \ \ \ \ \ \ \ \ \ 

% \textbf{result}=or(startwith(\textless m0\textgreater),endwith(\textless num\textgreater)) \\
\textbf{\ourname:}& ((\textless m0\textgreater)(.*))|((.*)([0-9]))&\textcolor{green!60!black}{or(startwith(\textless m0\textgreater),endwith(\textless num\textgreater))}\\
\bottomrule

\textbf{Example 2.} &\multicolumn{2}{l}{Query: \textit{lines with the string \textless M0$\textgreater$ ending in zero or more of a capital letter.}}\\
\hline
\textbf{Gold:} &
\textbf{(\textless m0\textgreater)(((.*)([\textless A-Z\textgreater]))*)} & \textbf{concat(\textless m0\textgreater,star(endwith(\textless cap\textgreater)))} \\
\textbf{Deep-Regex$^{MML}$:}& ((\textless m0\textgreater)(1,))|((.*)([A-Z])) &\textcolor{red}{or}(repeat\_least(\textless m0\textgreater,1),\textcolor{red}{endwith(\textless cap\textgreater)})\\
\textbf{SoftRegex:}& ((\textless m0\textgreater)(1,))\&((.*)([A-Z])) & and(repeat\_least(\textless m0\textgreater,1),\textcolor{red}{endwith(\textless cap\textgreater)})\\
\textbf{\ourname -w/o SC+CI:} &(.*(.*\textless m0\textgreater.*))* & star(endwith(\textcolor{red}{contain(\textless m0\textgreater)}))\\
\textbf{\ourname -w/o SC:}& (\textless m0\textgreater)(((.*)([\textless A-Z\textgreater]))*) &\textcolor{green!60!black}{concat(\textless m0\textgreater,star(endwith(\textless cap\textgreater)))}\\
% step1=endwith(\textless cap\textgreater) \\
% \ \ \ \ \ \ \ \ \ \ \ \ \ \ \ \ \ \ \ \ \ \ \ \ \ \  step2=star(step1) \\
% \ \ \ \ \ \ \ \ \ \ \ \ \ \ \ \ \ \ \ \ \ \ \ \ \ \  step3=concat(\textless m0\textgreater,step2) \\
% \ \ \ \ \ \ \ \ \ \ \ \ \ \ \ \ \ \ \ \ \ \ \ \ \ \ 
% \textbf{result}=concat(\textless m0\textgreater,star(endwith(\textless cap\textgreater))) \\
\textbf{\ourname:}&  (\textless m0\textgreater)(((.*)([\textless A-Z\textgreater]))*) &\textcolor{green!60!black}{concat(\textless m0\textgreater,star(endwith(\textless cap\textgreater)))}\\
\bottomrule

\textbf{Example 3.} &\multicolumn{2}{l}{Query: \textit{items with a any letter preceding \textless M0\textgreater.}}\\
\hline
\textbf{Gold:} &\textbf{(([A-Z])|([a-z]))(\textless m0\textgreater)} & \textbf{concat(or(\textless low\textgreater,\textless cap\textgreater),\textless m0\textgreater)} \\
\textbf{Deep-Regex$^{MML}$:}& (([a-z])\&([a-z] ))(\textless m0\textgreater) &concat(\textcolor{red}{and(\textless low\textgreater,\textless low\textgreater)},\textless m0\textgreater)\\
\textbf{SoftRegex:}& ([a-z])(\textless m0\textgreater) & concat(\textcolor{red}{\textless low\textgreater},\textless m0\textgreater)\\
\textbf{\ourname -w/o SC+CI:} & (([A-Z])\&([a-z]))(\textless m0\textgreater) & concat(\textcolor{red}{and}(\textless low\textgreater,\textless cap\textgreater),\textless m0\textgreater)\\
\textbf{\ourname -w/o SC:}& (([A-Z])\&(([a-z])\&(a-zA-Z)))(\textless m0\textgreater) &concat(\textcolor{red}{and(and}(\textless low\textgreater,\textless cap\textgreater),\textless let\textgreater),\textless m0\textgreater)\\
% step1=endwith(\textless cap\textgreater) \\
% \ \ \ \ \ \ \ \ \ \ \ \ \ \ \ \ \ \ \ \ \ \ \ \ \ \  step2=star(step1) \\
% \ \ \ \ \ \ \ \ \ \ \ \ \ \ \ \ \ \ \ \ \ \ \ \ \ \  step3=concat(\textless m0\textgreater,step2) \\
% \ \ \ \ \ \ \ \ \ \ \ \ \ \ \ \ \ \ \ \ \ \ \ \ \ \ 
% \textbf{result}=concat(\textless m0\textgreater,star(endwith(\textless cap\textgreater))) \\
\textbf{\ourname:}&  (([A-Z])|(([a-z])|(a-zA-Z)))(\textless m0\textgreater) &\textcolor{green!60!black}{concat(or(or(\textless low\textgreater,\textless cap\textgreater),\textless let\textgreater),\textless m0\textgreater)}\\
\bottomrule
\end{tabular}
}  %end setlength
% \vspace{-2mm}
\end{table*}

We further qualitatively analyze concrete regexes generated by various approaches. Three examples are listed in Table~\ref{table_examples}. 

\begin{itemize}
\item In Example 1, the query ``\texttt{either start with string \textless M0$\textgreater$ or end with any number}'' involves two steps, namely, ``\texttt{start with string \textless M0\textgreater}'' and ``\texttt{end with any number}'', which are successfully predicted by \ourname. %trained with chains of inference. 
By contrast, all language models without chain of inference mistakenly put the "or" operator in the middle, despite the correct operators and operands. %which indicates their apparent limitation in capturing the structural information of regexes. 
This suggests that the chain of inference captures the real text-matching processes of regexes better than plain regex and parenthetical notation for regex trees.
% and chain of inference makes the reasoning process of language model more interpretable, much like human experience.

\item Example 2 compares the generated regexes for the query ``\texttt{lines with the string \textless M0$\textgreater$ ending in zero or more of a capital letter}''. The query is clearly relevant to ``\texttt{capital letter}''. However, %\ourname with tree PLM mistakenly omits it, while
Deep-Regex$^{MML}$ and SoftRegex mistakenly omit the ``\texttt{zero or more}'' information.
Both \ourname and its variant without self-consistency correctly generate the gold regex, because the chain of inference decomposes the regex into multiple easier sub-regexes, and therefore helps the language model understand the complete information of the query.

\item Example 3 shows the results of ``\texttt{items with a any letter preceding \textless M0\textgreater}''. The ground truth is ``\texttt{or(\textless low\textgreater,\textless cap\textgreater)}'', while Deep-Regex$^{MML}$ and SoftRegex misunderstand ``\texttt{any letter}'' as any lower case letter, and the variants of \ourname misunderstand the keyword \texttt{letter} as ``\texttt{and(\textless low\textgreater,\textless cap\textgreater)}''. Although both variants generate the wrong regexes when decoding greedily, \ourname with self-consistency predicts the right answer. 
\end{itemize}

These examples demonstrate the superiority of \ourname in regex generation, affirming the effectiveness of chain of inference and self-consistency decoding.

\medskip\noindent\textbf{Answer to RQ1:} \ourname can achieve better DFA-EQ and EM accuracy than the state-of-the-art regex generation methods and popular tree generation method. Even if we replace the backbone of S$_2$RE with T5, \ourname achieves better performance. The advantage of \ourname is more obvious on the complex datatset \datasetb.

\subsection{Effect of Chain of Inference (RQ2)}

We conduct a number of ablation experiments to explore the effect of chain of inference.

 \subsubsection{Chain of inference in \ourname}
 
 We compare two variants to verify the
efficacy of chain of inference, namely,  \ourname without self-consistency (-w/o SC) and \ourname with neither self-consistency nor chain of inference (-w/o SC+CI). The former is built with a chain PLM that only generates chains of inference, while the latter is built with a PLM that directly generates trees in the form of parenthetical notated sequences. The results are shown in the bottom half of Table~\ref{table_rq1}.

As we can see, chain of inference, by step-by-step rationale, can greatly enhance the performance of regex generation. Compared with a simple PLM (-w/o SC+CI), \ourname without self-consistency (-w/o SC) gains visible improvements in most metrics. 
For example, the EM increases by about 2.4\% and 2.7\%  on \datasetb and \dataseta, respectively.

 \subsubsection{Chain of inference with different PLMs}
We further explore the generalizability of the chain of inference with different PLMs. Specifically, we train GPT2, BART-small, BART-base, T5-small, and T5-base on both (query, regex) and (query, chain) pairs, respectively, and evaluate their DFA-EQ@1 accuracy on the \datasetb dataset. The results are presented in Table~\ref{table_cio}. 
As the results show, most PLMs perform better when trained with chains of inference than trained with regexes in the form of parenthetical notated sequences.
The improvement from the chain of inference is most significant on GPT2, BART-small, and BART-base, which are about 7.1\%, 7.6\%, and 6.9\%. We conjecture that these models do not fully understand the regex in the format of parenthetical notated sequences, so the chain of inference can greatly increase their performance. Comparatively, the T5 model has achieved high performance on \datasetb, which obscures the improvement brought by the chain of inference.

The results corroborate previous findings that median-sized language models like T5-base can also solve certain multi-step problems through inference~\cite{ho2022large}, as compared to ultra-large language models such as ChatGPT.
Unlike using chain-of-thought prompting for large language models, we fine-tune T5 using a sufficient number of automatically generated chains of inference. Such task-specific fine-tuning can be seen as an effective way to stimulate the reasoning ability of T5-base on downstream tasks~\cite {ho2022large}.

\medskip\noindent\textbf{Answer to RQ2:} Chain of inference makes an appreciable contribution to the performance of \ourname in regex generation. In addition, chain of inference can lead to considerable performance improvements on different PLMs.

\begin{figure}[t!]
	\centering
	%0.237
    	\begin{subfigure}[b]{0.237\textwidth}
    		\centering
    		\includegraphics[width=\textwidth,trim=0 0 0 0, clip]{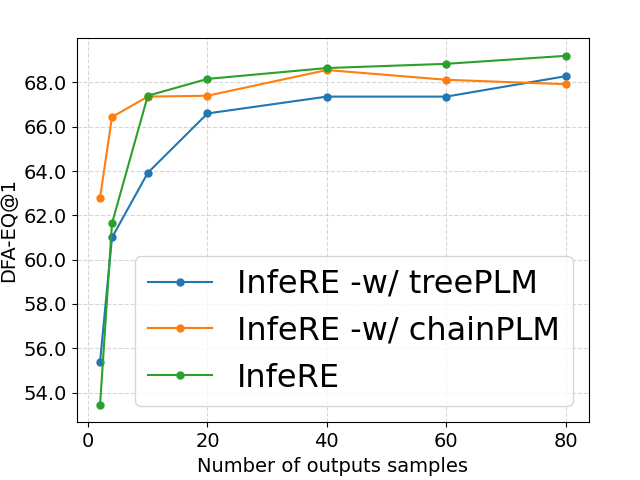}
    		\subcaption{\datasetb}
    		\vspace{2mm}
    		\label{sub1}
    	\end{subfigure}
    	\begin{subfigure}[b]{0.237\textwidth}
    		\centering
    		\includegraphics[width=\textwidth,trim=0 0 0 0,clip]{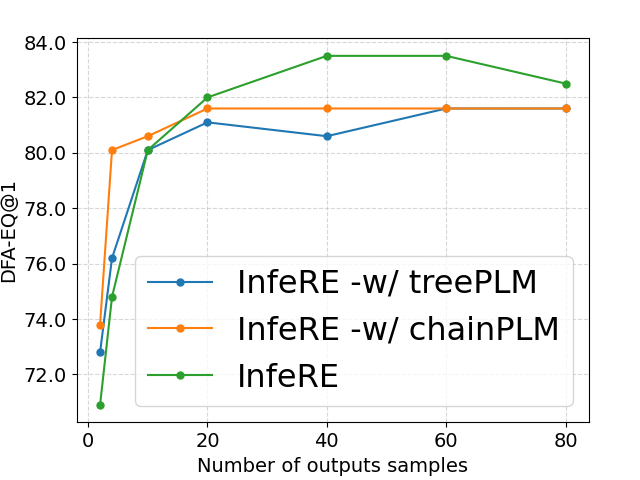}
    		\subcaption{\dataseta}
    		\vspace{2mm}
    		\label{sub2}
    	\end{subfigure}
    	\vspace{0mm}	
	\caption{Performance of \ourname under different numbers of sampled outputs. %with different PLMs on \dataseta and \datasetb datasets
	}
%	\vspace{-4mm}
	\label{fig:sc}
\end{figure}

\begin{table}[t]
\caption{Effectiveness of chain of inference with different PLMs. We evaluate  DFA-EQ@1(\%) on the \datasetb dataset.%Larger language models provide greater performance.
$w/o$ CI: the model trained on (query, regex) pairs; $w/$ CI: the model trained on (query, chain) pairs.} 
\label{table_cio}
\begin{center}
%\small
\setlength{\tabcolsep}{1.5mm}{
\begin{tabular}{l|ccccc}
 \toprule
\bf Approach & \bf GPT2 & \bf BART-small & \bf BART-base & \bf T5-small & \bf T5-base \\
\midrule
- $w/o$ CI   & 52.0 & 55.4 & 55.9 & 65.6 & 67.2\\
- $w/$ CI  & 55.7 & 59.6 & 59.8 & 65.2 & 67.8 \\
\bottomrule
\end{tabular}
}
%\vspace{-2mm}	
\end{center}
%\vspace{-4mm} 
\end{table}

\subsection{Effect of Self-Consistency Decoding (RQ3)}

%\subsubsection{Effectiveness of self-consistency decoding}

We conduct an ablation experiments to explore the effect of self-consistency decoding. Specially, we compare \ourname against \ourname without self-consistency (-w/o SC) to study the effect of self-consistency decoding. The results are shown in the bottom half of Table~\ref{table_rq1}.
 
As the result shows, self-consistency decoding demonstrates its effectiveness. 
\ourname significantly outperforms the variant without self-consistency (-w/o SC), with 1.6\% and 3.9\% improvement on average in terms of DFA-EQ@1 and DFA-EQ@5.
The improvement is more significant when self-consistency decoding and chain of inference mechanisms are used together, e.g., 4.4\% in terms of DFA-EQ@5 on \datasetb. 
%We can observe that all DFA-EQ accuracy drops when removing self-consistency decoding.

However, we have also observed that \ourname without self-consistency gains the best EM score. The reason for this, in addition to the chain of inference, is that self-consistency decoding takes semantic consistency as the metric. Hence it boosts the generation of more regexes with different syntax but the same semantics, which leads to a decrease in the EM accuracy. The performance degradation on EM metric does not affect the actual application, since we are more concerned with semantically correct regexes in regex generation task.

\textit{Impact of number of output samples:}
As the self-consistency decoding relies on the number $k$ of output samples, we vary $k$ and evaluate the  DFA-EQ accuracy of our approach on the two datasets.
Besides \ourname, which employs both a chain PLM and a tree PLM to perform self-consistency decoding, we also evaluate \ourname with only one single PLM, i.e., tree PLM or chain PLM. 
The results are shown in Figure~\ref{fig:sc}.

As the number of outputs sampled increases, all three models tend to perform better. The two variants with only single PLM achieve 68.6\% and 68.3\% DFA-equal accuracy on \datasetb relying on
self-consistency respectively, outperforming the result 67.8\% and 67.2\%
when decoding greedily. \ourname with multi PLMs achieve the best performance, 69.2\% DFA-equal accuracy. Specially, we can see that when $k$ exceeds 15, \ourname with multi PLMs consistently outperforms two variants with a single PLM. The reason is that self-consistency decoding aggregates the outputs from different PLMs and thus reduces the variance of the final results.

Besides, \ourname with only chain PLM achieves stronger performance than that with tree PLM under almost all numbers of output samples. The difference is more significant when $k$ is relatively small. This indicates that the chain of inference enhances the stability of the model in regex generation, which helps the self-consistency decoding to sample correct answer even when output samples are insufficient. 

\medskip\noindent\textbf{Answer to RQ3:} Self-consistency decoding can significantly help \ourname generate semantically correct regexes, achieving 1.6\% and 3.9\% improvement on average in terms of DFA-EQ@1 and DFA-EQ@5 metrics, and the effect of self-consistency decoding becomes much more apparent as the number of ouput samples increases.%\zs{Deleted}
% In addition, self-consistency decoding with multi PLMs consistently outperforms self-consistency decoding with single PLM for all sample numbers, which affirms the effectiveness of multi PLMs.\gu{too long answer}

\subsection{Impact of Different Backbone PLMs (RQ4)}
We further assess the impacts of different backbone PLMs on the performance of \ourname.
We experiment with two types of backbone models with two sizes on the \datasetb dataset, including BART-base, BART-small, T5-small, and T5-base.
The results are shown in Table~\ref{table_rq2}. 

Overall, T5 surpasses BART in the regex generation task. The pre-training tasks of BART include document-level denoising autoencoding. We conjecture that its pre-training objectives and pre-training data make it inappropriate for structured expression generation. In particular, T5-base performs the best in terms of all metrics. Clearly, larger language models boost greater performance in regex generation.  %Table ~\ref{table_backbonedfa} shows the DFA-EQ performance of \ourname 

\medskip\noindent\textbf{Answer to RQ4:} The type and size of backbone PLM strongly affects the performance of \ourname and \ourname with T5-base backbone performs best.

\begin{table}[t]
\caption{Ablation results of backbone PLMs on \datasetb. %Larger language models provide greater performance.
}
\label{table_rq2}
\begin{center}
%\small
\setlength{\tabcolsep}{1.5mm}{
\begin{tabular}{l|ccc}
 \toprule
\bf Model & \bf DFA-EQ@1(\%) & \bf DFA-EQ@5(\%) & \bf EM(\%) \\
\midrule
T5-base   & 69.2 & 89.3& 55.5\\
T5-small  & 66.8 & 88.9 & 51.7 \\
BART-base  & 59.4 & 79.9  & 43.6\\ 
BART-small  & 60.9 & 70.9& 42.3 \\ 
\bottomrule
\end{tabular}
}
%\vspace{-2mm}	
\end{center}
%\vspace{-4mm} 
\end{table}

%\begin{table}[t]
%\begin{center}
%\small
%\begin{tabular}{l|ccc}
% \toprule
%Model& \bf \makecell{\ourname \\-w/o(SC+CI)} & \bf \makecell{\ourname \\-w/o(SC)} & \bf \ourname \\
%\midrule
%T5-base   & 67.2\% & 67.8\%& 69.2\%\\
%T5-small  & 65.6\% & 65.2\% & 66.8\% \\
%BART-base  & 55.9\% & 59.8\%  & 59.4\%\\ 
%BART-small  & 55.4\% & 59.6\%& 60.9\% \\ 
%\bottomrule
%\end{tabular}
%\caption{DFA-EQ@1 of \ourname with different decoders on the \datasetb dataset varying backbone pre-trained models . %Larger language models provide greater performance. 
%}
%\label{table_backbonedfa}
%\end{center}
%\end{table}

% \begin{table}[t]
% \begin{center}
% \small
% \begin{tabular}{l|cccc}
%  \toprule
% Model Size & \bf BART-small & \bf BART-base & \bf T5-small & \bf T5-base \\
% \midrule
% \ourname   & 60.9\% & 59.4\%& 66.8\% & 69.2\%\\
% ~~- $w/o$ SC  & 59.6\% & 59.8\% & 65.2\% & 67.8\% \\
% ~~- $w/o$ CI  & 55.4\% & 55.9\%  & 65.6\% & 67.2\%\\ 

% \bottomrule
% \end{tabular}
% \caption{DFA-EQ@1 of \ourname with different decoders on the NL-RX-Turk dataset varing backbone pre-trained models . %Larger language models provide greater performance.
% }
% \label{table_backbonedfa}
% \end{center}
% \end{table}

\subsection{Impact of Data Size (RQ5)}
% \begin{figure}[t]
%     \centerline{\includegraphics[width=0.5\textwidth]{figures/Size.png}}
%     \caption{Performance of various training set sizes on the \datasetb dataset.}
%     \label{fig:size}
% %\vspace{-2mm} 
% \end{figure}

\begin{figure}[t!]
	\centering
	%0.237
    	\begin{subfigure}[b]{0.237\textwidth}
    		\centering
    		\includegraphics[width=\textwidth,trim=0 0 0 0, clip]{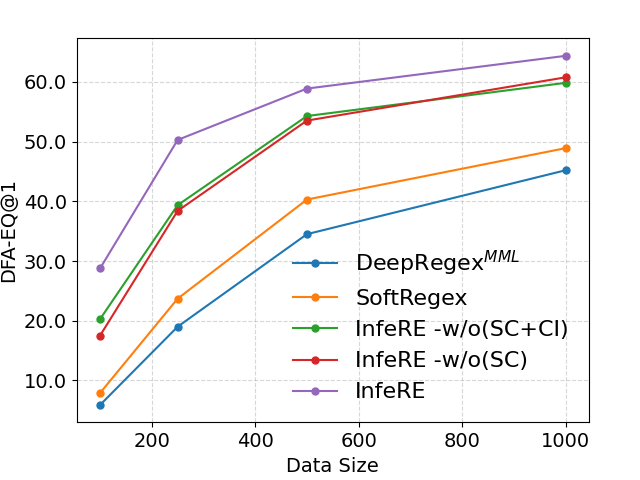}
    		\subcaption{Size from 100 to 1000}
    		\vspace{2mm}
    		\label{sub1}
    	\end{subfigure}
    	\begin{subfigure}[b]{0.237\textwidth}
    		\centering
    		\includegraphics[width=\textwidth,trim=0 0 0 0,clip]{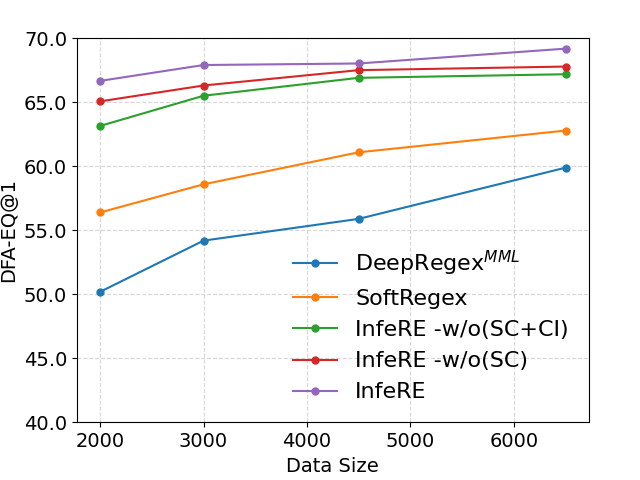}
    		\subcaption{Size from 2000 to 6500}
    		\vspace{2mm}
    		\label{sub2}
    	\end{subfigure}
	%\vspace{0mm}    	
	\caption{Performance under different training set sizes on \datasetb.
	}
	\label{fig:size}
	%\vspace{-3mm}
\end{figure}

We also study the effect of training data size on the \datasetb dataset. 
We vary the size of data from 100 to full data, i.e., 6500, and investigate its effect on the DFA-EQ accuracy of regex generation. Specifically, we compare \ourname, \ourname (-w/o SC), \ourname (-w/o SC+CI), SoftRegex and DeepRegex$^{MML}$. Figure~\ref{fig:size} shows the results.

As we can see, \ourname and its variants consistently outperform baseline models under different training sizes. 
Notably, when the training data gets smaller (e.g., $\textless $1000), the improvement becomes much more significant. This indicates that our approach has a strong ability in a low-resource generation.
We also observe a more profound advantage of the self-consistency decoding in the low-resource setting, as opposed to the chain of inference mechanism, which exhibits a more substantial improvement under conditions of sufficient data. This phenomenon may be attributed to the fact that the low-resource setting tends to result in less robust performance, which can be alleviated through integrated sampling via the self-consistency decoding mechanism. Conversely, the chain of inference mechanism is much more suitable for enabling a language model to learn structural information in situations where sufficient data is available.

\medskip\noindent\textbf{Answer to RQ5: } \ourname outperforms the baseline models under
all data sizes. In particular, self-consistency decoding exhibits a more significant advantage in low-resource settings, while the chain of inference contributes more to the performance when sufficient data is available.

\subsection{Threats to Validity}

We have identified the following threats to our approach:

%\subsubsection{Training Data} Although \ourname achieves superb performance on low-resource data compared with baseline methods, the chains of inference we propose struggle in limited training data. The chains of inference represent the structural information of regexes more obviously but bring additional output complexity, which means greater demand for training data resources. It remains to investigate how such chains of inference are more effective in scarce data.

%\subsubsection{Computational Cost} Another downside of \ourname is that the pre-trained model it takes leads to the requirement of large GPU resources and longer training time. Different from the traditional LSTM-based model, pre-trained model has larger parameters and a more complex network structure, bringing more time and computational cost. We leave more efficient training methods for future directions.

\subsubsection{The format of inference chains} In our experiments, we represent chain of inference as nodes of parsing trees. Results show that our chain of inference has little effect on the performance when the training data size is extremely limited. Hence it remains to investigate whether other formats of inference chains can lead to better performance.

\subsubsection{Generalization} 
We evaluate the efficacy of the chain of inference on two regex datasets. 
In theory, the chain of inference we propose can be extended to any tree-based generation task.  We leave the evaluation on other related tasks for future work.

%\section{Discussion}
%\label{sec:discuss}
%\subsection{How does chain of inference help regex generation?}

 % and eliminates the cost of manual annotation due to automatic script. Theoretically, our neural inference chain can be applied to any domain with tree structured outputs, regardless of the size of datasets. 
%Our method does not need to modify the loss function and parameters of language models or training and decoding procedures. 

%\subsection{Difference from few-shot learning}
%Our work differs from recent work on chain-of-thought prompting.  While they use chain-of-thought as prompts to elicit reasoning in large language models, our method automatically generates chains of internal processes and assembles them into the final regexes.

\section{Conclusion}
\label{sec:conclusion}

This paper introduces \ourname, a novel approach to generate regular expressions from natural language descriptions. \ourname formulates the process of regex generation as a chain of thought and leverages pre-trained language models to generate step-by-step internal processes. 
The internal processes chains are assembled into the final regexes using a self-consistency decoding mechanism, which combines the sampled outputs to enhance the robustness of the generated regexes. Our experimental results demonstrate that \ourname achieves remarkable improvements on two benchmark datasets.
% This paper presents \ourname, a novel paradigm for generating regexes from natural language descriptions. \ourname formulates regex generation as a chain of thought and generates the step-by-step internal processes of building regexes by leveraging pre-trained language models. The chains of internal processes are assembled into the final regexes through a self-consistency decoding mechanism that ensembles the sampled outputs to enhance the robustness. Experimental results show that \ourname achieves remarkable improvement on two benchmark datasets. %While we have evaluated this method on regexes generation, our technique is general and can be applied to other tasks with tree structured outputs. we leave other tasks related to our neural inference chain for future work. 
In the future, we will explore the chain of inference in other forms and apply it to more code intelligent tasks. For example, a program can be regarded as a chain of module interaction. The introduction of the chain of inference has the potential to improve the capability of existing models to generate a large complete program instead of function-level code. 

Source code and dataset to reproduce our work are available at
\href{https://github.com/Smallqqqq/InfeRE}{https://github.com/Smallqqqq/InfeRE}.

\section*{Acknowledgment}
This research is supported by National Natural Science Foundation of China (Grant No. 62232003, 62102244, 62032004 and 62272296) and CCF-Tencent Open Research Fund (RAGR20220129). 

\balance
\bibliographystyle{template/IEEEtran.bst}
\bibliography{main}

\begin{thebibliography}{10}
\providecommand{\url}[1]{#1}
\csname url@samestyle\endcsname
\providecommand{\newblock}{\relax}
\providecommand{\bibinfo}[2]{#2}
\providecommand{\BIBentrySTDinterwordspacing}{\spaceskip=0pt\relax}
\providecommand{\BIBentryALTinterwordstretchfactor}{4}
\providecommand{\BIBentryALTinterwordspacing}{\spaceskip=\fontdimen2\font plus
\BIBentryALTinterwordstretchfactor\fontdimen3\font minus
  \fontdimen4\font\relax}
\providecommand{\BIBforeignlanguage}[2]{{%
\expandafter\ifx\csname l@#1\endcsname\relax
\typeout{** WARNING: IEEEtran.bst: No hyphenation pattern has been}%
\typeout{** loaded for the language `#1'. Using the pattern for}%
\typeout{** the default language instead.}%
\else
\language=\csname l@#1\endcsname
\fi
#2}}
\providecommand{\BIBdecl}{\relax}
\BIBdecl

\bibitem{davis2019aren}
J.~C. Davis, L.~G. Michael~IV, C.~A. Coghlan, F.~Servant, and D.~Lee, ``Why
  aren’t regular expressions a lingua franca? an empirical study on the
  re-use and portability of regular expressions,'' in \emph{Proceedings of the
  2019 27th ACM Joint Meeting on European Software Engineering Conference and
  Symposium on the Foundations of Software Engineering}, 2019, pp. 443--454.

\bibitem{luo2018marrying}
B.~Luo, Y.~Feng, Z.~Wang, S.~Huang, R.~Yan, and D.~Zhao, ``Marrying up regular
  expressions with neural networks: A case study for spoken language
  understanding,'' \emph{arXiv preprint arXiv:1805.05588}, 2018.

\bibitem{Semantic-Unify}
N.~Kushman and R.~Barzilay, ``Using semantic unification to generate regular
  expressions from natural language,'' in \emph{Human Language Technologies:
  Conference of the North American Chapter of the Association of Computational
  Linguistics, Proceedings, June 9-14, 2013, Westin Peachtree Plaza Hotel,
  Atlanta, Georgia, {USA}}, 2013, pp. 826--836.

\bibitem{Deep-Regex}
N.~Locascio, K.~Narasimhan, E.~DeLeon, N.~Kushman, and R.~Barzilay, ``Neural
  generation of regular expressions from natural language with minimal domain
  knowledge,'' in \emph{Proceedings of the 2016 Conference on Empirical Methods
  in Natural Language Processing, {EMNLP} 2016, Austin, Texas, USA, November
  1-4, 2016}, 2016, pp. 1918--1923.

\bibitem{SemRegex}
Z.~Zhong, J.~Guo, W.~Yang, J.~Peng, T.~Xie, J.~Lou, T.~Liu, and D.~Zhang,
  ``Semregex: {A} semantics-based approach for generating regular expressions
  from natural language specifications,'' in \emph{Proceedings of the 2018
  Conference on Empirical Methods in Natural Language Processing, Brussels,
  Belgium, October 31 - November 4, 2018}, 2018, pp. 1608--1618.

\bibitem{SoftRegex}
J.~Park, S.~Ko, M.~Cognetta, and Y.~Han, ``Softregex: Generating regex from
  natural language descriptions using softened regex equivalence,'' in
  \emph{Proceedings of the 2019 Conference on Empirical Methods in Natural
  Language Processing and the 9th International Joint Conference on Natural
  Language Processing, {EMNLP-IJCNLP} 2019, Hong Kong, China, November 3-7,
  2019}, 2019, pp. 6424--6430.

\bibitem{ye2020benchmarking}
X.~Ye, Q.~Chen, I.~Dillig, and G.~Durrett, ``Benchmarking multimodal regex
  synthesis with complex structures,'' \emph{arXiv preprint arXiv:2005.00663},
  2020.

\bibitem{chen2020multi}
Q.~Chen, X.~Wang, X.~Ye, G.~Durrett, and I.~Dillig, ``Multi-modal synthesis of
  regular expressions,'' in \emph{Proceedings of the 41st ACM SIGPLAN
  conference on programming language design and implementation}, 2020, pp.
  487--502.

\bibitem{ye2020optimal}
X.~Ye, Q.~Chen, I.~Dillig, and G.~Durrett, ``Optimal neural program synthesis
  from multimodal specifications,'' \emph{arXiv preprint arXiv:2010.01678},
  2020.

\bibitem{DeepSketch}
X.~Ye, Q.~Chen, X.~Wang, I.~Dillig, and G.~Durrett, ``Sketch-driven regular
  expression generation from natural language and examples,'' \emph{Trans.
  Assoc. Comput. Linguistics}, vol.~8, pp. 679--694, 2020.

\bibitem{TransRegex}
Y.~Li, S.~Li, Z.~Xu, J.~Cao, Z.~Chen, Y.~Hu, H.~Chen, and S.~Cheung,
  ``{TransRegex:} multi-modal regular expression synthesis by
  generate-and-repair,'' in \emph{43rd {IEEE/ACM} International Conference on
  Software Engineering, {ICSE} 2021, Madrid, Spain, 22-30 May 2021}.\hskip 1em
  plus 0.5em minus 0.4em\relax {IEEE}, 2021, pp. 1210--1222.

\bibitem{kojima2022large}
T.~Kojima, S.~S. Gu, M.~Reid, Y.~Matsuo, and Y.~Iwasawa, ``Large language
  models are zero-shot reasoners,'' \emph{arXiv preprint arXiv:2205.11916},
  2022.

\bibitem{wei2022chain}
J.~Wei, X.~Wang, D.~Schuurmans, M.~Bosma, E.~Chi, Q.~Le, and D.~Zhou, ``Chain
  of thought prompting elicits reasoning in large language models,''
  \emph{arXiv preprint arXiv:2201.11903}, 2022.

\bibitem{DIN-SQL}
M.~Pourreza and D.~Rafiei, ``Din-sql: Decomposed in-context learning of
  text-to-sql with self-correction,'' \emph{arXiv preprint arXiv:2304.11015},
  2023.

\bibitem{wang2022self}
X.~Wang, J.~Wei, D.~Schuurmans, Q.~Le, E.~Chi, and D.~Zhou, ``Self-consistency
  improves chain of thought reasoning in language models,'' \emph{arXiv
  preprint arXiv:2203.11171}, 2022.

\bibitem{spishak2012type}
E.~Spishak, W.~Dietl, and M.~D. Ernst, ``A type system for regular
  expressions,'' in \emph{Proceedings of the 14th Workshop on Formal Techniques
  for Java-like Programs}, 2012, pp. 20--26.

\bibitem{liu2019lightweight}
X.~Liu, Y.~Jiang, and D.~Wu, ``A lightweight framework for regular expression
  verification,'' in \emph{2019 IEEE 19th International Symposium on High
  Assurance Systems Engineering (HASE)}.\hskip 1em plus 0.5em minus 0.4em\relax
  IEEE, 2019, pp. 1--8.

\bibitem{AlphaRegex}
M.~Lee, S.~So, and H.~Oh, ``Synthesizing regular expressions from examples for
  introductory automata assignments,'' in \emph{Proceedings of the 2016 ACM
  SIGPLAN International Conference on Generative Programming: Concepts and
  Experiences}, 2016, pp. 70--80.

\bibitem{RegexGenerator}
A.~Bartoli, A.~De~Lorenzo, E.~Medvet, and F.~Tarlao, ``Inference of regular
  expressions for text extraction from examples,'' \emph{IEEE Transactions on
  Knowledge and Data Engineering}, vol.~28, no.~5, pp. 1217--1230, 2016.

\bibitem{li2020flashregex}
Y.~Li, Z.~Xu, J.~Cao, H.~Chen, T.~Ge, S.-C. Cheung, and H.~Zhao, ``Flashregex:
  deducing anti-redos regexes from examples,'' in \emph{2020 35th IEEE/ACM
  International Conference on Automated Software Engineering (ASE)}.\hskip 1em
  plus 0.5em minus 0.4em\relax IEEE, 2020, pp. 659--671.

\bibitem{XML1}
G.~J. Bex, F.~Neven, T.~Schwentick, and S.~Vansummeren, ``Inference of concise
  regular expressions and dtds,'' \emph{ACM Transactions on Database Systems
  (TODS)}, vol.~35, no.~2, pp. 1--47, 2010.

\bibitem{XML2}
G.~J. Bex, W.~Gelade, F.~Neven, and S.~Vansummeren, ``Learning deterministic
  regular expressions for the inference of schemas from xml data,'' \emph{ACM
  Transactions on the Web (TWEB)}, vol.~4, no.~4, pp. 1--32, 2010.

\bibitem{XML3}
D.~D. Freydenberger and T.~K{\"o}tzing, ``Fast learning of restricted regular
  expressions and dtds,'' in \emph{Proceedings of the 16th International
  Conference on Database Theory}, 2013, pp. 45--56.

\bibitem{Ranta1998}
A.~Ranta, ``A multilingual natural-language interface to regular expressions,''
  in \emph{International Workshop on Finite State Methods in Natural Language
  Processing}, 1998, p. 79–90.

\bibitem{narang2020wt5}
S.~Narang, C.~Raffel, K.~Lee, A.~Roberts, N.~Fiedel, and K.~Malkan, ``Wt5?!
  training text-to-text models to explain their predictions,'' \emph{arXiv
  preprint arXiv:2004.14546}, 2020.

\bibitem{wiegreffe2021reframing}
S.~Wiegreffe, J.~Hessel, S.~Swayamdipta, M.~Riedl, and Y.~Choi, ``Reframing
  human-ai collaboration for generating free-text explanations,'' \emph{arXiv
  preprint arXiv:2112.08674}, 2021.

\bibitem{wang2022rationale}
X.~Wang, J.~Wei, D.~Schuurmans, Q.~Le, E.~Chi, and D.~Zhou,
  ``Rationale-augmented ensembles in language models,'' \emph{arXiv preprint
  arXiv:2207.00747}, 2022.

\bibitem{ye2022unreliability}
X.~Ye and G.~Durrett, ``The unreliability of explanations in few-shot
  in-context learning,'' \emph{arXiv preprint arXiv:2205.03401}, 2022.

\bibitem{Transformer}
A.~Vaswani, N.~Shazeer, N.~Parmar, J.~Uszkoreit, L.~Jones, A.~N. Gomez,
  L.~Kaiser, and I.~Polosukhin, ``Attention is all you need,'' in
  \emph{Advances in Neural Information Processing Systems 30: Annual Conference
  on Neural Information Processing Systems}, 2017, pp. 5998--6008.

\bibitem{yin2018tranx}
P.~Yin and G.~Neubig, ``Tranx: A transition-based neural abstract syntax parser
  for semantic parsing and code generation,'' \emph{arXiv preprint
  arXiv:1810.02720}, 2018.

\bibitem{T5}
C.~Raffel, N.~Shazeer, A.~Roberts, K.~Lee, S.~Narang, M.~Matena, Y.~Zhou,
  W.~Li, and P.~J. Liu, ``Exploring the limits of transfer learning with a
  unified text-to-text transformer,'' \emph{J. Mach. Learn. Res.}, vol.~21, pp.
  140:1--140:67, 2020.

\bibitem{stanovich2000individual}
K.~E. Stanovich and R.~F. West, ``Individual differences in reasoning:
  Implications for the rationality debate?'' \emph{Behavioral and brain
  sciences}, vol.~23, no.~5, pp. 645--665, 2000.

\bibitem{williams1992simple}
R.~J. Williams, ``Simple statistical gradient-following algorithms for
  connectionist reinforcement learning,'' \emph{Machine learning}, vol.~8,
  no.~3, pp. 229--256, 1992.

\bibitem{jiang2021ast}
H.~Jiang, L.~Song, Y.~Ge, F.~Meng, J.~Yao, and J.~Su, ``An ast structure
  enhanced decoder for code generation,'' \emph{IEEE/ACM Transactions on Audio,
  Speech, and Language Processing}, vol.~30, pp. 468--476, 2021.

\bibitem{dong2022antecedent}
Y.~Dong, G.~Li, and Z.~Jin, ``Antecedent predictions are dominant for
  tree-based code generation,'' \emph{arXiv preprint arXiv:2208.09998}, 2022.

\bibitem{beau2022impact}
N.~Beau and B.~Crabb{\'e}, ``The impact of lexical and grammatical processing
  on generating code from natural language,'' \emph{arXiv preprint
  arXiv:2202.13972}, 2022.

\bibitem{shen2022incorporating}
S.~Shen, X.~Zhu, Y.~Dong, Q.~Guo, Y.~Zhen, and G.~Li, ``Incorporating domain
  knowledge through task augmentation for front-end javascript code
  generation,'' in \emph{Proceedings of the 30th ACM Joint European Software
  Engineering Conference and Symposium on the Foundations of Software
  Engineering}, 2022, pp. 1533--1543.

\bibitem{ho2022large}
N.~Ho, L.~Schmid, and S.-Y. Yun, ``Large language models are reasoning
  teachers,'' \emph{arXiv preprint arXiv:2212.10071}, 2022.

\end{thebibliography}

\end{document}